\documentclass{article}
\usepackage{color}
\usepackage{amssymb}
\usepackage[cmex10]{amsmath}
\newtheorem{theorem}{Theorem}
\newtheorem{lemma}{Lemma}
\newtheorem{definition}{Definition}

\newtheorem{remark}{Remark}
\usepackage{spconf,amsmath,epsfig}
\usepackage{algorithm,algorithmic}
\usepackage{threeparttable}
\usepackage{tabularx,booktabs,graphicx,epsfig,subfigure}
\usepackage{float}
\usepackage{multirow}
\renewcommand{\algorithmicrequire}{ \textbf{Input:}}
\renewcommand{\algorithmicensure}{ \textbf{Output:}}
\newcommand{\tabincell}[2]{\begin{tabular}{@{}#1@{}}#2\end{tabular}}%
\begin{document}\sloppy

\def\x{{\mathbf x}}
\def\L{{\cal L}}

\title{Efficient Two-Dimensional Sparse Coding Using Tensor-Linear Combination}
%
\name{Fei Jiang$^1$, Xiao-Yang Liu$^{1,2}$, Hongtao Lu$^{1}$, Ruimin Shen$^{1}$}
\address{$^1$Department of Computer Science and Engineering, Shanghai Jiao Tong University\\
         $^2$Department of Electrical Engineering, Columbia University}

\address{}
\maketitle

\begin{abstract}
Sparse coding (SC) is an automatic feature extraction and selection technique that is widely used in unsupervised learning.
However, conventional SC vectorizes the input images, which breaks apart the local proximity of pixels and destructs the elementary object structures of images.
In this paper, we propose a novel two-dimensional sparse coding (2DSC) scheme that represents the input images as the tensor-linear combinations under a novel algebraic framework. 2DSC learns much more concise dictionaries because it uses the circular convolution operator, since the shifted versions of atoms learned by conventional SC are treated as the same ones.  We apply 2DSC to natural images and demonstrate that 2DSC returns meaningful dictionaries for large patches. Moreover, for mutli-spectral images denoising, the proposed 2DSC reduces computational costs with competitive performance in comparison with the state-of-the-art algorithms.
\end{abstract}
\section{Introduction}
\label{sec:intro}

Sparse coding (SC) is a classical unsupervised feature extraction technique for finding concise representations of the data, which has been successfully applied to numerous areas across computer vision and pattern recognition \cite{wright2010sparse}.  Conventional SC \cite{lee2006efficient} aims to approximate vector-valued inputs by linear combinations of a few bases. Such bases correspond to patterns that physically represent elementary objects, and they compose a dictionary.

Conventional SC \cite{lee2006efficient} model suffers from the following two major problems: 1) the vectorization preprocess elementarily breaks apart the local proximity and destructs the object structures of images; and 2) the high computational complexity restricts its applications, thus small sizes of patches are usually used. Usually, the dictionary is overcomplete as the number of bases is larger than the dimension of the input image data. Therefore, the dictionary size is significantly large for high dimensional data, meaning that it requires prohibitive large number of computations for conventional SC.

However, existing approaches cannot solve the above two problems satisfactorily. Two kinds of SC models are proposed to preserve the spatial proximity of images, which are tensor sparse coding (TenSR) \cite{qi2016tensr,qi2013two} and convolutional sparse coding (CSC) \cite{bristow2013fast,heide2015fast}. For TenSR models \cite{qi2016tensr,qi2013two}, a series of separable dictionaries are adopted to approximate the structures in each mode of the input data. Though the sizes of the dictionaries are significantly reduced, the relationships among the modes are ignored. The object structures usually distribute across all modes of the data. For CSC models \cite{bristow2013fast,heide2015fast}, the dictionaries are used to capture local patterns, and the convolution operator is introduced to learn the shifting-invariant patterns. However, optimizing such models with the convolution operator are computational challenging. Moreover, each feature map (sparse representation) has nearly the same size as the input image, which is quite larger than conventional SC, and will increase the resources for storage and the computational complexity.
\begin{figure}[!t]
\centering
\includegraphics[height=0.175\textwidth]{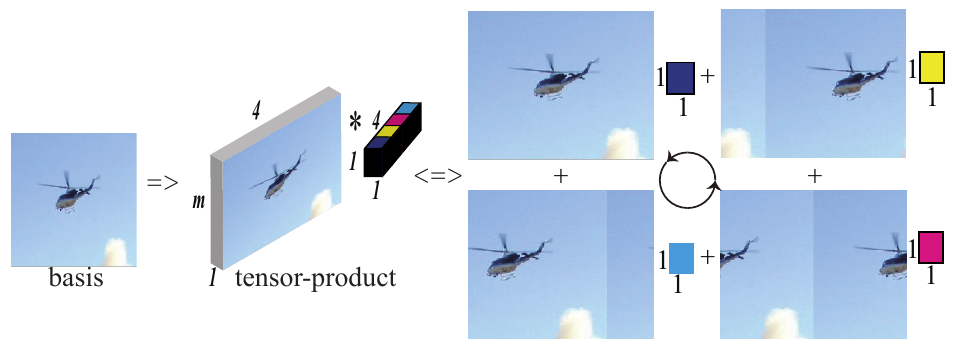}
\caption{Shifting invariance involved in tensor-product. The shifted versions correspond to a dynamic flight in a counter-clockwise direction.  \label{fig1}}
\end{figure}

The basic idea that motivates us to address these challenges lies in two aspects: 1) tensor representation is able to preserve the local proximity and to capture the elementary object structures; and 2) we exploit the tensor-product operation under a novel algebraic framework where the tensor-linear combinations are used to approximate the images, instead of the tucker decomposition used in TenSR \cite{qi2016tensr,qi2013two} and convolution operation in CSC \cite{bristow2013fast,heide2015fast}. For one aspect, tensor-product based on circular convolution operation, which can generate the data by shifted versions of bases. Fig. \ref{fig1} shows the shifted versions generated from the tensor-product without storing them.  For another aspect, the tensor-linear combination (see Definition \ref{tlindef}) is a generalization of the standard linear combination. The number of required bases  can be significantly reduced, which also reduces the computational complexity.

In this paper, we propose a novel sparse coding model, two-dimensional sparse coding (2DSC), in which the input images are represented as third-order tensors, and the tensor-linear combinations are used for data approximation. To solve the 2DSC problem, a novel alternating minimization algorithm is presented which includes a sparse coding learning step and a dictionary learning step. For sparse coding, we propose a new iterative shrinkage thresholding algorithm based on tensor-product, which is directly implemented in the tensor space. For dictionary learning, we show that it can be solved efficiently by transforming to a Lagrange dual problem in the frequency domain.

The rest of this paper is organized as follows: Section \ref{sec:2} introduces the notations and preliminary used in our paper. Section \ref{sec:3} presents the proposed 2DSC model, followed by the novelties of our model. In Section \ref{sec:4}, an efficient alternating minimization algorithm for 2DSC is proposed.  We demonstrate the effectiveness of our 2DSC model by simulation experiment and multi-spectral images denoising in Section \ref{sec:5}. Finally, we conclude the paper in Section \ref{sec:6}.
To summarize, this paper makes the following contributions:
\begin{itemize}
  \item We propose a novel two-dimensional sparse coding model (2DSC) for image representation, which preserves the local proximity of pixels and elementary object structures of images. 2DSC is superior in dealing with high dimensional data.
  \item We discuss the geometric properties of the dictionary returned by 2DSC, and show that there exists an equivalent sum space spanned by the corresponding vectorizations of the bases. Therefore, the dictionary of 2DSC has a stronger representation capability than that of the conventional SC.
  \item We propose an efficient alternating minimization algorithm. For coefficient learning, a novel iterative shrinkage thresholding algorithm based on tensor-product is provided which exploits the optimization of mechanism in the tensor space. For dictionary learning, we convert it into a corresponding problem in the frequency domain and solve the dictionary by dual Lagrange  which significantly reduces the number of variables to be optimized.
\end{itemize}

\section{Related Work}
In this section, we briefly review the related work on sparse coding models with consideration of spatial structures of the images, including sparse coding based on tensor representations \cite{qi2016tensr,qi2013two} and convolutional sparse coding \cite{bristow2013fast,heide2015fast}.

The first stream is the tensor-based sparse coding (TenSR) models \cite{qi2016tensr,qi2013two}, which preserves the spatial structures of images by tensor representations. The data is represented based on tucker decomposition. Instead of using one dictionary as conventional SC, a series of separable dictionaries are adopted to model data, and each dictionary is corresponding to one dimension of the data. Therefore, the sizes of the dictionaries are significantly reduced due to the small size of each dimension compared with the sizes of the data. Though the sizes of dictionaries are significantly reduced, the relationships between the modes are ignored. The object structures are usually distributed across all dimensions of the data.

The second stream is the convolutional sparse coding (CSC) models \cite{bristow2013fast,heide2015fast}, which represents an image as the summation of convolutions of the feature maps and the corresponding filters. Convolution operator models the similarities of the proximal pixels, which further preserves the spatial structures of images. However, optimizing CSC models are computational challenging. Moreover, each feature map (sparse representation) has nearly the same size as the image, which is quite larger than conventional SC, and will increase the resources for storage and the computational complexity.
\begin{figure*}[!t]
\centering
\includegraphics[height=0.1\textwidth]{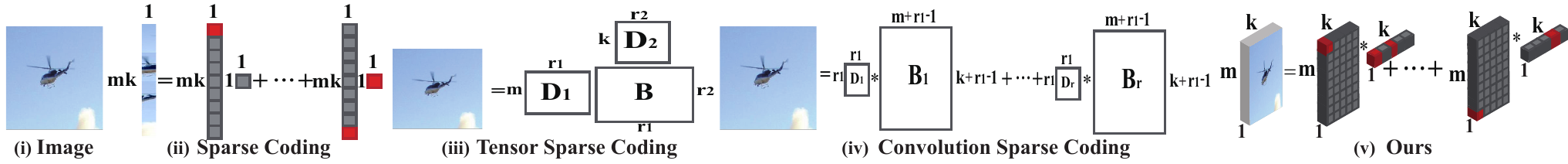}
\caption{Sparse Coding Models. (i) an $m\times k$ image, (ii) SC is based on linear combination, where $mk$ bases can represent the $\mathbb{R}^{mk}$ space. (iii) TenSR \cite{qi2016tensr} is based on tucker decomposition and (iv) CSC \cite{heide2015fast} is based on convolution operator, which are not easy to determine the number of bases in their spaces. (v) Our model is based on tensor-linear combination, where $m$ tensor bases can represent the $\mathbb{K}^m$ tensor space. \label{fig2}}
\end{figure*}
\section{Notation and Preliminary}
\label{sec:2}
A third-order tensor is denoted as $\mathcal{X}\in\mathbb{R}^{m\times n \times k}$.  The expansion of $\mathcal{X}$ along the third dimension is represented as $\underline{\mathcal{X}}=[\mathcal{X}^{(1)};\mathcal{X}^{(2)};\cdots;\mathcal{X}^{(k)}]$. The transpose of tensor $\mathcal{X}$ is denoted as $\mathcal{X}^{\dagger}$, where $\mathcal{X}^{\dagger(1)}=\mathcal{X}^{(1)^T}$, and $\mathcal{X}^{\dagger(\ell)}=\mathcal{X}^{(k+2-\ell)^T}$, $2\leq\ell\leq k$, and the superscript ``T" represents the transpose of matrices. The discrete Fourier transform (DFT) along the third dimension of $\mathcal{X}$ is denoted as $\widehat{\mathcal{X}}$.

For convenience, tensor spaces $\mathbb{R}^{1\times1\times k}$, $\mathbb{R}^{m\times1\times k}$, and $\mathbb{R}^{m\times n\times k}$ are denoted as $\mathbb{K}$, $\mathbb{K}^m$, and $\mathbb{K}^{m\times n}$, respectively. $[k]$ denotes the set $\{1,2,\cdots,k\}$. The $\ell_1$ and Frobenius norms of tensors are denoted as $\|\mathcal{X}\|_{1}=\sum_{i,j,\ell}|\mathcal{X}(i,j,\ell)|$, and $\|\mathcal{X}\|_F=\left(\sum_{i,j,\ell}\mathcal{X}(i,j,\ell)^2\right)^{1/2}$. Furthermore, we need the following definitions.
\begin{definition}\cite{kilmer2013third}
The tensor-product between $\mathcal{D}\in\mathbb{K}^{m\times r}$ and $\mathcal{B}\in\mathbb{K}^{r\times n}$ is a tensor $\mathcal{X}\in\mathbb{K}^{m\times n}$ where $\mathcal{X}(i,j,:) = \sum_{q=1}^{r}\mathcal{D}(i,q,:)*\mathcal{B}(q,j,:)$, and $*$ denotes the circular convolution operation.
\end{definition}
\begin{remark}
The tensor-product $\mathcal{X}=\mathcal{D}*\mathcal{B}$ can be efficiently computed in the frequency domain as:
\begin{equation}
\label{f-domain}
\widehat{\mathcal{X}}^{(\ell)} = \widehat{\mathcal{D}}^{(\ell)}\widehat{\mathcal{B}}^{(\ell)},~\ell\in[k].
\end{equation}
\end{remark}
\begin{lemma}{\cite{kilmer2013third}}
\label{lemma1}
The tensor-product $\mathcal{X}=\mathcal{D}*\mathcal{B}$ has an equivalent matrix-product as:
\begin{equation}
\label{t-prod}
\underline{\mathcal{X}} = \underline{D}^c\underline{\mathcal{B}},
\end{equation}
where $\underline{D}^c$ is the circular matrix of $\mathcal{D}$ defined as follows:
\begin{equation}
\label{cirmat}
 {\underline{D}}^c = \left(
               \begin{array}{cccc}
                 \mathcal{D}^{(1)} & \mathcal{D}^{(k)} & \cdots & \mathcal{D}^{(2)} \\
                 \mathcal{D}^{(2)} & \mathcal{D}^{(1)} & \cdots & \cdots \\
                 \cdots & \cdots & \cdots & \mathcal{D}^{(k)} \\
                 \mathcal{D}^{(k)}& \mathcal{D}^{(k-1)} & \cdots &\mathcal{D}^{(1)}\\
               \end{array}
             \right).
\end{equation}
\end{lemma}
\begin{definition}\cite{kilmer2013third}
The tensor-linear combinations of the tensor bases $\{\mathcal{D}_j\}_{j=1}^{r}\subset\mathbb{K}^{m}$ with the corresponding tensor coefficient $\{\mathcal{B}_j\}_{j=1}^r\subset\mathbb{K}$ are defined as:
\begin{equation}
\label{tlindef}
\mathcal{D}_1*\mathcal{B}_1 + \cdots + \mathcal{D}_r*\mathcal{B}_r=\mathcal{D}*\mathcal{B}.
\end{equation}
\end{definition}
where $\mathcal{D}\in\mathbb{K}^{m\times r}$ with $\mathcal{D}(:,j,:)=\mathcal{D}_j$, and $\mathcal{B}\in\mathbb{K}^r$ with $\mathcal{B}(j,1,:)=\mathcal{B}_j$.
\begin{remark}
The tensor-linear combination is a generalization of the linear combination.
\end{remark}
\begin{definition}
\label{def4}
The spanned tensor space $\mathbb{W}$  by the tensor bases set of $\mathcal{D}\in\mathbb{K}^{m\times r}$  is defined as $\mathbb{W}=\left\{\mathcal{X}\in\mathbb{K}^m|\mathcal{X}=\mathcal{D}*\mathcal{B},\mathcal{B}\in\mathbb{K}^r\right\}$.
\end{definition}
\section{Problem Statement}
\label{sec:3}
\subsection{Problem Formulation}
Instead of preprocessing images into vectors, we represent $n$ images of size $m\times k$ by $\mathcal{X}\in\mathbb{K}^{m\times n}$, and propose a novel sparse coding model, named two-dimensional sparse coding (2DSC), as follows:
\begin{eqnarray}
\label{2DSCobj}
 \min_{\mathcal{D},\mathcal{B}} && \frac{1}{2}\|\mathcal{X} - \mathcal{D} * \mathcal{B}\|_F^2 + \beta\|\mathcal{B}\|_{1}\nonumber\\
 \rm{s.t.} && \|\mathcal{D}(:,j,:)\|_F^2 \leq 1,\quad j\in[r],
\end{eqnarray}
where  $\mathcal{D}\in\mathbb{K}^{m\times r}$ is the tensor dictionary where each lateral slice $\mathcal{D}(:,j,:)$ is a basis, $\mathcal{B}\in\mathbb{K}^{r\times n}$ is the tensor coefficient. The parameter $\beta$ balances the approximation error and the sparsity of the tensor coefficients, and $r$ is the number of atoms. Conventional SC is a special case of (\ref{2DSCobj}) when $k=1$.
\subsection{Novelties of 2DSC}
The proposed 2DSC is not a simply extension of conventional SC on the two-dimensional data, which has novel properties. The first one is the size of dictionary in 2DSC can be significantly reduced without damaging the reconstruction accuracy due to the tensor-linear combination. The second one is shifting invariant which means that the data can be generated from 2DSC model by the shifted versions of bases without explicitly storing them.
\subsubsection{``Slim" Dictionary}
\begin{lemma}\cite{kilmer2013third}
\label{lemma2}
Tensor space $\mathbb{K}^{m}$ can be generated by $m$ tensors from an orthogonal set.
\end{lemma}
Fig. \ref{fig2} shows four kinds of sparse coding models for an $m\times k$ image representation.  In SC \cite{lee2006efficient}, $\mathbb{R}^{mk}$ space is generated by $mk$ bases based on linear combination, while in 2DSC, $\mathbb{K}^m$ can be spanned with only $m$ bases. In TenSR \cite{qi2016tensr} and CSC \cite{heide2015fast}, it is not easy to determine the number of bases for the image space. Though CSC \cite{heide2015fast} is also based on convolution operator, the sparse representations are nearly the size as the input images, which are quite larger than ours.
Lemma \ref{lemma2} means an $m\times k$ data can be generated by only $m$ elements with the same size based on tensor linear combination. However,  $mk$ elements are required based on linear combinations. With much fewer atoms for data representation can significantly reduce the computational complexity, which shows the potential applications of 2DSC model in high dimensional data.

\subsubsection{Shifting Invariance}
\begin{theorem}
\label{th1}
The tensor space spanned by $\mathcal{D}\in\mathbb{K}^{m\times r}$ defined in (\ref{def4}) is equivalent to a sum space of $k$ vector subspaces in $\mathcal{R}^{mk}$.
\end{theorem}

\begin{figure}[!t]
\centering
\includegraphics[height=0.13\textwidth]{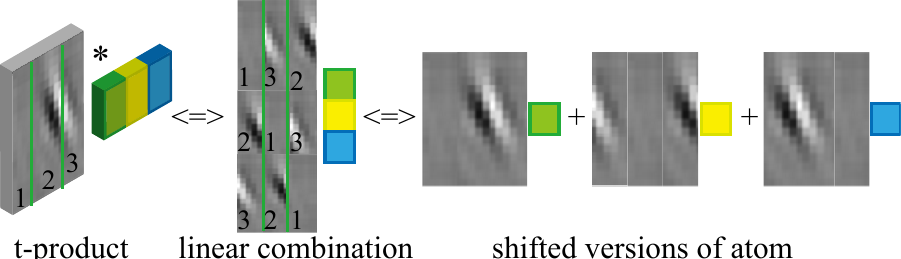}
\caption{The leftmost tensor production is equivalent to the linear combination in middle one, which corresponds to the shifted versions in rightmost one.\label{fig3} }
\end{figure}

As shown in (\ref{t-prod}), (\ref{tlindef}) is equivalent to $\underline{\mathcal{D}}^c\underline{B}\in\mathbb{R}^{mk}$ in the vector space which is actually a sum space as following:
\begin{equation}
\label{eq:t-lin-eqv}
\begin{split}
\underline{\mathcal{D}}^c\underline{\mathcal{B}}
                        &=[\mathbf{D},\mathbf{D}_1,\cdots,\mathbf{D}_{k-1}][b;b_1;\cdots;b_{k-1}]\\
                        &=\mathbf{D}b+\mathbf{D}_1b_1+\cdots+\mathbf{D}_{k-1}b_{k-1}.
\end{split}
\end{equation}
where $\mathbf{D}$ is the vectorizations of $\mathcal{D}$, the bases in $\mathbf{D}_j$ are circular shifted versions of those in $\mathbf{D}$. $b$ and $b_j$ are the coefficients to the corresponding dictionaries $\mathbf{D}$  and $\mathcal{D}_j$.   Fig. \ref{fig1}  explicitly shows the shifted versions of a base generated  by tensor-product if the image of helicopter is seen as a basis.
The space generated by the tensor linear combination can be transferred to a sum space of vector spaces generated by the linear combination, which includes the shifted versions of the original atoms.
Fig. \ref{fig3} explicitly shows the equivalent linear combination and the shifted versions of a $m\times 3$ atom generated from tensor product by its twisted form of size $m\times1\times3$ and a $1\times1\times 3$ coefficient.
\section{Alternating Minimization Algorithm}
\label{sec:4}
Problem (\ref{2DSCobj}) is quite challenging due to the non-convex objective function and the convolution operator. Instead of transforming (\ref{2DSCobj}) into conventional SC formulation based on Lemma \ref{lemma1}, we propose an efficient algorithm by alternately optimizing $\mathcal{B}$ and $\mathcal{D}$ in the tensor space, as shown in Algorithm \ref{alg1}.
\begin{algorithm}[t!]
\caption{Algorithm for 2DSC \label{alg1}}
\begin{algorithmic}[1]
 \renewcommand{\algorithmicrequire}{\textbf{Input:}}
 \renewcommand{\algorithmicensure}{\textbf{Output:}}
 \REQUIRE $n$ images: $\mathcal{X}\in\mathbb{K}^{m\times n}$, the number of atoms: $r$, sparsity regularizer: $\beta>0$, maximum iterative steps: num,
 \noindent\STATE{\textbf{Initialization:}  Randomly initialize $ \mathcal{D}\in\mathbb{K}^{m\times r}$, $\mathcal{B}:=0\in\mathbb{K}^{r\times n}$, and Lagrange dual variables $\lambda\in\mathbb{R}^r$,}
 \FOR{$\text{iter}=1$ to num}
   \STATE{//\textbf{Tensor Sparse Coding Learning}}\\
   \STATE{Solving $\mathcal{B}$ via Equation (\ref{Sobj3}) in Algorithm \ref{alg2},}
   \STATE{//\textbf{Tensor Dictionary Learning}}\\
   \STATE{$\widehat{\mathcal{X}}=\text{fft}(\mathcal{X},[~],3)$, $\widehat{\mathcal{B}}=\text{fft}(\mathcal{B},[~],3)$,}
   \FOR{$\ell=1$ to $k$}
    \STATE{Solving (\ref{dual}) for $\Lambda$ by Newton's method,}
    \STATE{Calculate $\widehat{\mathcal{D}}^{(\ell)}$ from (\ref{dic}),}
   \ENDFOR
   \STATE{$\mathcal{D} = \text{ifft}(\widehat{\mathcal{D}},[~],3)$},
 \ENDFOR
\ENSURE $\mathcal{D}$, $\mathcal{B}$.
\end{algorithmic}
\end{algorithm}

\subsection{Learning Tensor Coefficient}
For clarity, we discuss how to solve the tensor sparse representation for an image of size $m\times k$, which is represented as $\mathcal{X}\in\mathbb{K}^m$.

Given the dictionary $\mathcal{D}\in\mathbb{K}^{m\times r}$, solving the tensor sparse representations of images $\mathcal{B}\in\mathbb{K}^r$ are converted to the following problem as:
\begin{equation}
\label{Sobj}
\min_{\mathcal{S}\in\mathbb{K}^{r}}\frac{1}{2}\|\mathcal{X}-\mathcal{D}*\mathcal{B}\|_F^2 + \beta\|\mathcal{B}\|_{1}.
\end{equation}
By Lemma \ref{lemma1}, (\ref{Sobj}) can be solved by conventional sparse coding algorithms, which is equivalent to
\begin{equation}
\label{Sobj1}
\min_{\mathbf{b}\in\mathbb{R}^{rk}}\frac{1}{2}\|\mathbf{x} - \mathbf{D}\mathbf{b}\|_F^2 + \beta\|\mathbf{b}\|_{1}
\end{equation}
where $\mathbf{x}=\underline{\mathcal{X}}\in\mathbb{R}^{mk}$, $\mathbf{D}=\underline{\mathcal{D}}^c\in\mathbb{R}^{mk\times rk}$, and $\mathbf{b}=\underline{\mathcal{B}}\in\mathbb{R}^{rk}$.
The size of the dictionary $\mathbf{D}$ in (\ref{Sobj1}) is significantly increased with the size of images, which also increase the computational complexity.

To alleviate this problem, we propose a novel Iterative Shrinkage Thresholding algorithm based on Tensor-product (ISTA-T) to solve (\ref{Sobj}) directly. We first rewrite (\ref{Sobj}) as:
\begin{equation}
\label{Sobj2}
\min_{\mathcal{B}\in\mathbb{K}^{r\times n}} f(\mathcal{B}) + \beta g(\mathcal{B}),
\end{equation}
where $f(\mathcal{B})$ stands for the data reconstruction term $\frac{1}{2}\|\mathcal{X}-\mathcal{D}*\mathcal{B}\|_F^2$ and $g(\mathcal{B})$ stands for the sparsity constraint term $\|\mathcal{B}\|_{1}$. An iterative shrinkage algorithm is used to solve (\ref{Sobj2}), which can  be rewritten as a linearized function around the previous estimation $\mathcal{B}_{p}$ with the proximal regularization and the non-smooth regularization. Thus, at the $p+1$- th iteration, $\mathcal{B}_{p+1}$ can be updated by
\begin{equation}
\label{Sobj3}
\begin{aligned}
\mathcal{B}_{p+1} = \arg\min_{\mathcal{B}}&f(\mathcal{B}_{p}) + \langle\nabla f(\mathcal{B}_{p}),\mathcal{B}-\mathcal{B}_{p}\rangle\\
&+ \frac{L_{p+1}}{2}\|\mathcal{B}-\mathcal{B}_{p}\|_F^2 + \beta g(\mathcal{B}),
\end{aligned}
\end{equation}
where $L_{p+1}$ is a Lipschitz constant, and $\nabla f(\mathcal{B})$ is the gradient defined in the tensor space. Then, (\ref{Sobj3}) is equivalent to
\begin{equation}
\label{Sobj4}
\mathcal{B}_{p+1} = \arg\min_{\mathcal{B}} \frac{1}{2}\|\mathcal{B} - (\mathcal{B}_{p}-\frac{1}{L_{p+1}}\nabla f(\mathcal{B}_{p}))\|_F^2 + \frac{\beta}{L_{p+1}}\|\mathcal{B}\|_{1}.
\end{equation}
To solve (\ref{Sobj4}), we firstly show $\nabla f(\mathcal{B})$ w.r.t. the data reconstruction term $\frac{1}{2}\|\mathcal{X}-\mathcal{D}*\mathcal{B}\|_F^2$:
\begin{equation}
\label{Sgrad}
\nabla f(\mathcal{B}) = \mathcal{D}^{\dagger}*\mathcal{D}*\mathcal{B} - \mathcal{D}^{\dagger}*\mathcal{X}.
\end{equation}
Secondly, we discuss how to determine the Lipschitz constant $L_{p+1}$ in (\ref{Sobj4}). For every $\mathcal{B}$, $\mathcal{C}$, we have
\begin{equation}
\label{Lconst}
\begin{aligned}
\|\nabla f(\mathcal{B}) - \nabla f(\mathcal{C})\|_F 
&=\|\underline{(\mathcal{D}^{\dagger}*\mathcal{D})}^c(\underline{\mathcal{B}}-\underline{\mathcal{C}})\|_F\\
&=\|\underline{(\mathcal{D}^{\dagger}*\mathcal{D})}^c(\underline{\mathcal{B}}-\underline{\mathcal{C}})\|_2\\
&\leq\|\underline{(\mathcal{D}^{\dagger}*\mathcal{D})}^c\|_2\|\underline{\mathcal{B}}-\underline{\mathcal{C}}\|_2\\
&\leq\|\underline{(\mathcal{D}^{\dagger}*\mathcal{D})}^c\|_F\|\mathcal{B} - \mathcal{C}\|_F\\
&=\sum_{\ell=1}^k\|\hat{\mathcal{D}}^{(\ell)^H}\hat{\mathcal{D}}^{(\ell)}\|_F^2\|\mathcal{B} - \mathcal{C}\|_F,
\end{aligned}
\end{equation}
where the superscript ``$H$" represents conjugate transpose.

Thus the Lipschitz constant of $f(\mathcal{B})$ used in our algorithm is $L(f)=\sum_{\ell=1}^k\|\widehat{\mathcal{D}}^{(\ell)^H}\widehat{\mathcal{D}}^{(\ell)}\|_F^2$.

Lastly, (\ref{Sobj4}) can be solved by the proximal operator $\textbf{Prox}_{\beta/L_{p+1}}(\mathcal{B}_{p}-\frac{1}{L_{p+1}}\nabla f(\mathcal{B}_{p}))$, where $\textbf{Prox}_{\tau}$ is the soft-thresholding operator $\textbf{Prox}_{\tau}(\cdot)\to \text{sign}(\cdot)\max(|\cdot| - \tau,0)$.

To speed up the convergence of the proposed ISTA-T, an extrapolation operator is adopted \cite{xu2014fast}. Algorithm \ref{alg2} summarizes the proposed ISTA-T algorithm.
\begin{algorithm}[t!]
\caption{Iterative Shrinkage Thresholding Algorithm based on Tensor-production \label{alg2}}
\begin{algorithmic}[1]
\REQUIRE $n$ images $\mathcal{X}\in\mathbb{K}^{m\times n}$, dictionary $\mathcal{D}\in\mathbb{K}^{m\times r}$, and the maximum iterations: num,
\STATE{\textbf{Initialization:} $\mathcal{B}_0:=0\in\mathbb{K}^{r\times n}$, Set $\mathcal{C}_1=\mathcal{B}_0\in\mathbb{K}^r$, $t_1=1$,}
\FOR {$p$ = 1 To \text{num} }
     \STATE{Set $L^p = \eta^p(\sum_{\ell=1}^k\|\widehat{\mathcal{D}}^{(\ell)^H}\widehat{\mathcal{D}}^{\ell}\|_F)$,}
     \STATE{Compute $\nabla f(\mathcal{C}_p)$ via Equation (\ref{Sgrad}),}
     \STATE{Compute $\mathcal{B}_p$ via $\textbf{Prox}_{\beta/L_p}(\mathcal{C}_p-\frac{1}{L_p}\nabla f(\mathcal{C}_p))$,}
     \STATE{$t_{p+1}=\frac{1+\sqrt{1+4t^2_p}}{2}$,}
     \STATE{$\mathcal{C}_{p+1}=\mathcal{B}_p + \frac{t_p-1}{t_{p+1}}(\mathcal{B}_p-\mathcal{B}_{p-1})$,}
\ENDFOR
\ENSURE Sparse Coefficient $\mathcal{B}$.
\end{algorithmic}
\end{algorithm}
\begin{table}[!t]
\centering
\caption{Complexity Analysis and Memory Usage of Sparse Represenation (SR) and Dictionary Learning (DL) for TenSR \cite{qi2016tensr} and Ours. \label{tab1}}
\scalebox{0.63}{\begin{tabular}{|c|c|c|c|c|}
\hline
\multicolumn{2}{|c|}{}      & Operation               & Complexity in Detail & Complexity\\
\hline
\multirow{2}{*}{SR}& TenSR \cite{qi2016tensr} & $\nabla f(\mathcal{B})$ & \tabincell{c}{$O(r_1^2r_2+r_2^2r_1+r_1^2m$\\$+r_2^2k+r_1r_2k+r_1mk)$} & $O(r_1^2r_2+r_1r_2^2)$\\
\cline{2-5}
                   & Ours                     &$\mathcal{D}^{\dagger}*\mathcal{D}*\mathcal{B}-\mathcal{D}^{\dagger}*\mathcal{X}$ & $O(3k\log k+3rmk)$ &  $O(rmk+k\log k)$\\
\hline
\multirow{3}{*}{DL}& \multirow{2}{*}{TenSR \cite{qi2016tensr}} & $\mathcal{A}_{(i)}$  &  $O(r_1r_2kn+r_1r_2mn)$ &\multirow{2}{*}{\tabincell{c}{$O(r_1^2kn+r_2^2mn+$\\$r_1r_2kn+r_1r_2mn)$}}       \\
\cline{3-4}
                   &                                           &$\min_{D_i}\|\mathcal{X}_{i}-D_i\mathcal{A}_{(i)}\|_F^2$ &$O(r_1^2kn+r_2^2mn)$ &                        \\
\cline{2-5}
                    & Ours  &$\min_{\mathcal{D}}\|\mathcal{X}-\mathcal{D}*\mathcal{B}\|_F^2$& $O(r^3+r^2n+rmn)$                               &  $O(r^2n)$     \\
\hline
\end{tabular}}
\end{table}
\subsection{Tensor Dictionary learning}
For learning the dictionary $\mathcal{D}$ while fixed $\mathcal{B}$, the optimization problem is:
\begin{eqnarray}
\label{2DSC_D}
\min_{\mathcal{D}\in\mathbb{K}^{m\times r}} && \frac{1}{2}\|\mathcal{X}-\mathcal{D}*\mathcal{B}\|_F^2\nonumber\\
\rm{s.t.} && \|\mathcal{D}(:,j,:)\|_F^2\leq 1, ~ j\in[r].
\end{eqnarray}
where atoms are coupled together due to the circular convolution operator. Therefore, we firstly decompose (\ref{2DSC_D}) into $k$ nearly-independent problems (that are coupled only through the norm constraint) by DFT as follows:
\begin{eqnarray}
\label{basis2}
 \min_{\widehat{\mathcal{D}}^{(\ell)},\ell\in[k]} && \sum_{\ell=1}^{k}\|\widehat{\mathcal{X}}^{(\ell)}-\widehat{\mathcal{D}}^{(\ell)}\widehat{\mathcal{B}}^{(\ell)}\|_F^2\nonumber\\
 \rm{s.t}. && \sum_{\ell=1}^{k}\|\widehat{\mathcal{D}}^{(\ell)}(:,j)\|_F^2\leq k, ~ j\in[r]
\end{eqnarray}

Then, we adopt the Lagrange dual \cite{lee2006efficient} for solving (\ref{basis2}) in frequency domain. The advantage of Lagrange dual is that the number of optimization variables is $r$, which is much smaller than $mkr$ in the primal problem (\ref{2DSC_D}).

To use the Lagrange dual algorithm, firstly, we consider the Lagrangian of (\ref{basis2}):
\begin{eqnarray}
 \mathcal{L}(\widehat{\mathcal{D}},\Lambda)&=&\sum_{\ell=1}^k\|\widehat{\mathcal{X}}^{(\ell)}-\widehat{\mathcal{D}}^{(\ell)}\widehat{\mathcal{B}}^{(\ell)}\|_F^2+\nonumber\\
 &&\sum_{j=1}^r\lambda_j\left(\sum_{\ell=1}^k\|\widehat{\mathcal{D}}^{(\ell)}(:,j)\|_F^2-k\right),
\end{eqnarray}
where $\lambda_j\geq0$, $j\in[r]$ is a dual variable, and $\Lambda=diag(\lambda)$.

Secondly, minimizing over $\widehat{\mathcal{D}}$ analytically, we obtain the optimal formulation of $\widehat{\mathcal{D}}$:
\begin{equation}
\label{dic}
\widehat{\mathcal{D}}^{(\ell)} = \left(\widehat{\mathcal{X}}^{(\ell)}\widehat{\mathcal{B}}^{(\ell)^H}\right)\left(\widehat{\mathcal{B}}^{(\ell)}\widehat{\mathcal{B}}^{(\ell)^H}+\Lambda\right)^{-1}, \ell\in[k].
\end{equation}
Substituting (\ref{dic}) into the Lagrangian $\mathcal{L}(\widehat{\mathcal{D}},\Lambda)$, we obtain the Lagrange dual function $\mathcal{D}(\Lambda)$:
\begin{equation}
\label{dual}
\mathcal{D}(\Lambda)=-\sum_{\ell=1}^k\text{Tr}\left(\widehat{\mathcal{B}}^{(\ell) ^H}\widehat{\mathcal{X}}^{(\ell)}\widehat{\mathcal{S}}^{(\ell)^H}\right)-k\sum_{j=1}^r\lambda_j,
\end{equation}
which is solved by Newton's method. Once getting the dual variables, the dictionary can be recovered by Equation (\ref{dic}).
\subsection{Complexity Analysis}
Table \ref{tab1} shows the computational complexity of our proposed 2DSC and TenSR \cite{qi2016tensr}.  The definitions of $\nabla f(\mathcal{B})$ and $\mathcal{A}_{i}$ can be found in \cite{qi2016tensr}. $r\geq\min\{m,k\}$, $r_1\geq m$ and $r_2\geq k$ are the sizes of dictionaries in our model and TenSR \cite{qi2016tensr}. The computational complexity of TenSR \cite{qi2016tensr} is higher than ours.  As shown in \cite{qi2016tensr}, for 21168 patches of size $5\times5\times5$, TenSR  takes 189 seconds for tensor coefficient learning, while in our model, it only takes about 23 seconds. The reason is due to the faster computation of tensor-product in frequency domain, which divides the original large size problems into much smaller ones.
\section{Evaluation}
\label{sec:5}
\subsection{Dictionaries for Large Patches}
\begin{figure}
\centering
\includegraphics[height=0.128\textwidth]{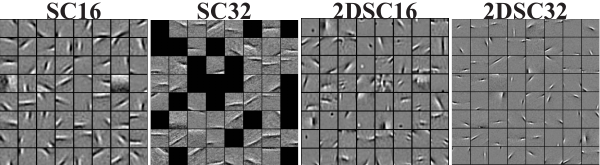}
\caption{Dictionaries learned from SC and 2DSC. For $16\times16$ patches, both SC and 2DSC can learn Gabor-like features, while for $32\times32$ patches, only 2DSC can learn meaningful features.  \label{fig5} }
\end{figure}
We analyze the learned dictionaries from conventional SC and 2DSC with the same bases and other parameters settings. Patches of sizes $d\times d$ ($d=16, 32$) extracted from Natural Images data are normalized to have zero means. The sparsity regularizers $\beta$ is set to 0.1, and the number of bases are 64, which makes sure the dictionary for 2DSC is overcompleted but not for SC. Fig. \ref{fig5} shows the learned dictionaries from 2DSC and conventional SC with different sizes of patches ($d=16, 32$). For small size  of patches ($d=16$), both models can learn the meaningful dictionaries (Gabor-like featues), such as edges, corners. For large size patches ($d=32$), the performances of SC are significantly degenerated, even the bases are becoming sparsity. But 2DSC still can learn  meaningful features.
\begin{table*}[!t]
\centering
\caption{PSNR and SSIM results of the different methods for different noise levels on balloons.\label{tab2}}
 \scalebox{0.95}{ \begin{tabular}{|c|c|c|c|c|c|c|c|c|c|c|}
  \hline
  \multirow{2}{*}{Method} & \multicolumn{5}{c|}{PSNR} & \multicolumn{5}{c|}{SSIM}\\
  \cline{2-11}
              &$\sigma=5$ & $\sigma=10$ &$\sigma=20$ &$\sigma=30$ &$\sigma=50$ &$\sigma=5$ &$\sigma=10$ &$\sigma=20$ &$\sigma=30$ &$\sigma=50$ \\
  \hline
  Noisy Image &34.16      &28.13        &22.11       &18.59       &14.15       & 0.8177    &0.5664      &0.2998      & 0.1912     &0.0983       \\
  BwK-SVD\cite{elad2006image}     &36.92      & 32.72       &28.90       &26.83       &24.21       & 0.9107    &0.8242      & 0.7136     &0.6487      &0.5617           \\
  3DK-SVD\cite{elad2006image}     &38.60      & 34.96       & 31.28      &29.00       & 26.15      & 0.9388    &0.8903      & 0.8238     &0.7775      &0.7261           \\
  LRTA \cite{renard2008}       &\textbf{42.98}      & \textbf{39.33}  &33.20       &32.99       &30.23       & 0.9664    &0.9424      & 0.9096     & 0.8622     &0.7904           \\
  PARAFAC \cite{liu2012}     &36.20      & 33.96       &32.96       &32.11       &27.49       &0.9230     &0.9099      &0.8555      &0.7897      &0.5636           \\
  TenSR \cite{qi2016tensr}      & 41.34     & 36.55       &32.24       & 31.25      &30.67       &0.9700    & 0.9283      &0.8623      & 0.8032     &0.7768           \\
  2DSC(Ours) &41.32       &36.56        & \textbf{33.21}  & \textbf{33.09}&\textbf{34.46}& \textbf{0.9766}&\textbf{0.9466}& \textbf{0.9274}&\textbf{0.8690}&\textbf{0.9248}          \\
  \hline
  \end{tabular}}
\end{table*}
\begin{figure}[!t]
\centering
\includegraphics[height=0.24\textwidth]{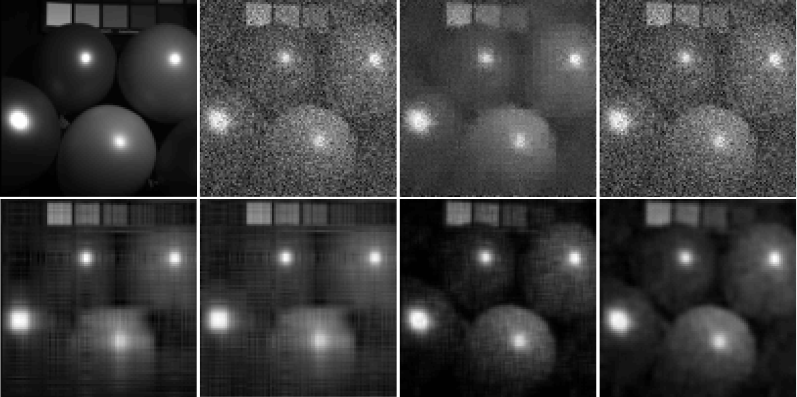}
\caption{Visual comparison of reconstruction results by different methods on ``balloons" in dataset \cite{yasuma2010}. From left to right: original image at 420nm band, noisy image, BwK-SVD\cite{elad2006image}, PARAFAC \cite{liu2012}, LRTA \cite{renard2008}, TenSR \cite{qi2016tensr}, and Ours.  \label{figd} }
\end{figure}
\subsection{Multi-spectral Image Denoising}
We apply 2DSC on multispectral images -- Columbia MSI Database \cite{yasuma2010}. Each dataset contains 31 real-world images of size $512\times 512$ and is collected from $400nm$ to $700nm$ at $10nm$ steps. We firstly to scale these images to [0, 255], and then add Gaussian white noise at different noise levels $\sigma=[5, 10, 20, 30, 50]$. In our 2DSC model, we extract $5\times5\times5$ patches of size from each noisy multi-spectral image, and save each patch into a tensor of size $25\times1\times5$. Dictionary of size $25\times30\times5$ are randomly initialized and trained iteratively ($\leq30$ iterations). Then we use the learned dictionaries to denoise the MSI images. Parameters in our scheme are $\beta = 10, 28, 200, 220, 290$, for $\sigma = 5, 10, 20, 30, 50$, respectively.

Table \ref{tab2} shows the comparison results in terms of  peak signal-to-noise ratio (PSNR) and structure similarity (SSIM) \cite{wang2004image} . There are 5 state-of-the-art MSI denoising methods are involved, including band-wise KSVD (BwK-SVD) method \cite{elad2006image}, 3D-cube KSVD (3DK-SVD) method \cite{elad2006image}, LRTA \cite{renard2008}, PARAFAC \cite{liu2012}, and TenSR \cite{qi2016tensr}. As shown in Table \ref{tab2}, our 2DSC outperforms all the comparison algorithms for the evaluations by SSIM, which measures the structure consistency between the target image and the reference image. For PSNR, our 2DSC works much better on higher level noises, especially for $\sigma=30$ and $\sigma=50$. For lower noise levels, LRTA \cite{renard2008} is the best one, ours and TenSR \cite{qi2016tensr} are comparable.

We add CSC \cite{heide2015fast} in the visualization of the denoising performances. Due to the high memory required by CSC \cite{heide2015fast} for the high resolution images, we resize the spectral images of balloons into $101\times101$, and only considering the first five bands. Fig. \ref{figd} shows the denoising results with the noise level $\sigma=30$. It is easy to observe that our method achieves the best denoising results. Note that for CSC \cite{heide2015fast}, we denoise each images separately which does not consider the information along the brands of spectral images. Filters of size $5\times5\times100$ are used, and the sparsity parameters are adjusted between [0.1, 10]. We guess the reason for the denoising results of CSC \cite{heide2015fast} may come from the small size of images. Due to the powerful representations of convolution operation, filters learn much noise information.
\section{Conclusion}
\label{sec:6}
In this paper, we propose a novel tensor based sparse coding algorithm, which can learn an efficient tensor representations of images by much smaller size of dictionary compared with conventional SC. Moreover, an much more efficient algorithm for the tensor sparse coefficients learning in tensor space is proposed. The effectiveness of our model has been demonstrated by dictionary learning for large sizes of patches. A following up work \cite{Jiang2017ICME} applies the 2DSC scheme to image clustering by incorporating a graph regularizer.

\bibliographystyle{IEEEbib}
\bibliography{TenSC}

\end{document}